\documentclass[conference, a4paper, 10pt]{IEEEtran}

\usepackage{amsmath}
\usepackage{amssymb}
\usepackage{amsthm}
\usepackage{hyperref}
\usepackage{tikz}
\usepackage{cite}
\usepackage{microtype}  
\usepackage{graphicx}
\usepackage{url}
\usepackage{booktabs}
\usepackage{multirow}
\usepackage{color}
\usepackage{xspace}
\usepackage[caption=false,font=footnotesize]{subfig}
\usepackage{pgfplots}
\pgfplotsset{compat=1.18}

\usepackage{tcolorbox}
\newtcolorbox{emquote}{
  colback=gray!10,
  colframe=black!50,
  boxrule=0.5pt,
  arc=2pt,
  left=8pt,
  right=8pt,
  top=4pt,
  bottom=4pt,
  fonttitle=\bfseries,
}

\newtheorem{theorem}{Theorem}
\newtheorem{definition}{Definition}

\title{Self-Organizing Survival Manifolds: A Theory for Unsupervised Discovery of Prognostic Structures in Biological Systems}

\author{
    \IEEEauthorblockN{\large Atahan Karagöz}
    \IEEEauthorblockA{
        \textit{Department of Computer Science} \\
        \textit{University of Basel} \\
        Basel, Switzerland \\
        atahan.karagoez@stud.unibas.ch
    }
}

\begin{document}

\maketitle

\begin{abstract}
    Survival is traditionally modeled as a supervised learning task, reliant on curated outcome labels and fixed covariates. This work rejects that premise. It proposes that survival is not an externally annotated target but a geometric consequence: an emergent property of the curvature and flow inherent in biological state space. We develop a theory of \textit{Self-Organizing Survival Manifolds} (SOSM), in which survival-relevant dynamics arise from low-curvature geodesic flows on latent manifolds shaped by internal biological constraints. A survival energy functional based on geodesic curvature minimization is introduced and shown to induce structures where prognosis aligns with geometric flow stability. We derive discrete and continuous formulations of the objective and prove theoretical results demonstrating the emergence and convergence of survival-aligned trajectories under biologically plausible conditions. The framework draws connections to thermodynamic efficiency, entropy flow, Ricci curvature, and optimal transport, grounding survival modeling in physical law. Health, disease, aging, and death are reframed as geometric phase transitions in the manifold’s structure. This theory offers a universal, label-free foundation for modeling survival as a property of form, not annotation—bridging machine learning, biophysics, and the geometry of life itself.
\end{abstract}    
\vspace{1em}
\begin{IEEEkeywords}
    Survival Analysis, Manifold Learning, Thermodynamics, Unsupervised Learning, Ricci Flow, Optimal Transport, Biological Systems, Phase Transitions, Entropy Minimization, Geometric Modeling
\end{IEEEkeywords}    

\section{Introduction}

Predicting survival trajectories remains a central challenge at the intersection of biology, medicine, and population-level inquiry. In clinical oncology, models of patient survival guide therapeutic decisions, stratify clinical trial cohorts, and inform personalized treatment planning. In epidemiology, survival modeling guides projections of disease burden, healthcare resource allocation, and public health strategies. Beyond disease, survival dynamics characterize biological processes such as aging, immunosenescence, and evolutionary adaptation.

Yet despite this centrality, survival modeling remains largely grounded in supervised learning paradigms. Classical methods such as Cox proportional hazards regression~\cite{cox1972regression} and its modern extensions require meticulously curated event-time annotations that encode the occurrence and timing of outcomes such as death, relapse, or systemic failure. While powerful, these approaches remain constrained by the limitations of survival annotations, which are frequently incomplete, noisy, or altogether absent—and often subject to cohort-induced selection bias in emerging biobank-scale repositories~\cite{oexner2025benchmarking}.

Meanwhile, advances in molecular profiling technologies have dramatically expanded the biological features available for prognosis. High-throughput transcriptomics, epigenomics, proteomics, and metabolomics yield high-dimensional molecular characterizations that reflect diverse axes of biological regulation. These "multi-omics" datasets capture the dynamic interplay of genomic, epigenetic, and post-transcriptional regulation underlying disease progression. However, the complexity and dimensionality of these datasets frequently surpass the statistical tractability and inferential robustness of conventional supervised models, especially when survival labels are sparse.

In parallel, unsupervised representation learning has emerged as a transformative approach to modeling complex high-dimensional data without reliance on annotations. Methods based on contrastive learning~\cite{chen2020simple}, manifold learning~\cite{tenenbaum2000global}, and self-supervised objectives~\cite{jing2019self} have demonstrated the ability to recover semantically meaningful structures purely from intrinsic data relationships. A representative instantiation of this principle is OmicsCL~\cite{karagoz2025omicsclunsupervisedcontrastivelearning}, which imposes survival-sensitive constraints within a contrastive formulation to reveal structure inherently aligned with differential survival. This lends credence to a deeper hypothesis: that survival is not merely annotated atop biological configurations, but instead arises from the latent geometry of the state space itself.

This intuition motivates a general geometric theory of survival. We propose that survival outcomes across biological systems are governed by the intrinsic geometry of latent biological manifolds. Specifically, we introduce the concept of \textit{Self-Organizing Survival Manifolds} (SOSM), in which patient trajectories align along low-curvature geodesic flows on the manifold, reflecting principles of minimal energy expenditure and entropy production.

This work makes the following contributions:

\begin{itemize}
    \item We introduce a universal geometric framework—Self-Organizing Survival Manifolds (SOSM)—to model survival dynamics without supervision.
    \item We derive a new energy-based loss functional, grounded in geodesic curvature minimization, and analyze its theoretical properties.
    \item We establish analytical connections between SOSM and foundational constructs from thermodynamics, Ricci flow, and optimal transport, situating survival within a geometric framework for biological inference.
    \item We propose a philosophical synthesis where health, disease, and aging emerge as geometric phases of biological manifolds.
\end{itemize}

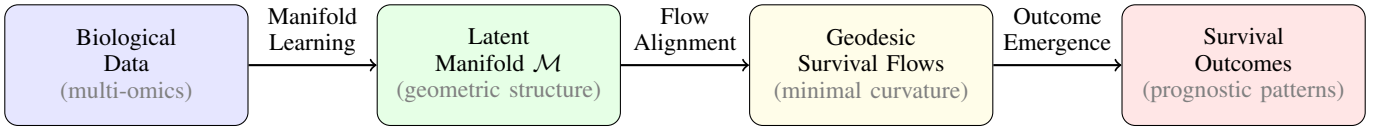
\begin{figure*}[ht]
    \centering
    \begin{tikzpicture}[node distance=4.9cm, auto, every node/.style={align=center, font=\small}]
        \tikzstyle{input}=[rectangle, draw, rounded corners=5pt, minimum width=3.2cm, minimum height=1.6cm, fill=blue!10]
        \tikzstyle{process}=[rectangle, draw, rounded corners=5pt, minimum width=3.2cm, minimum height=1.6cm, fill=green!10]
        \tikzstyle{flow}=[rectangle, draw, rounded corners=5pt, minimum width=3.2cm, minimum height=1.6cm, fill=yellow!10]
        \tikzstyle{output}=[rectangle, draw, rounded corners=5pt, minimum width=3.2cm, minimum height=1.6cm, fill=red!10]
    
        \node[input] (data) {Biological\\Data\\{\small \textcolor{gray}{(multi-omics)}}};
        \node[process, right of=data] (manifold) {Latent\\Manifold $\mathcal{M}$\\{\small \textcolor{gray}{(geometric structure)}}};
        \node[flow, right of=manifold] (flows) {Geodesic\\Survival Flows\\{\small \textcolor{gray}{(minimal curvature)}}};
        \node[output, right of=flows] (outcome) {Survival\\Outcomes\\{\small \textcolor{gray}{(prognostic patterns)}}};
    
        \draw[->, thick] (data.east) to[out=0, in=180] node[midway, above] {Manifold\\Learning} (manifold.west);
        \draw[->, thick] (manifold.east) to[out=0, in=180] node[midway, above] {Flow\\Alignment} (flows.west);
        \draw[->, thick] (flows.east) to[out=0, in=180] node[midway, above] {Outcome\\Emergence} (outcome.west);
    \end{tikzpicture}
    \caption{Schematic representation of the Self-Organizing Survival Manifolds (SOSM) framework. Biological data are mapped onto a latent manifold $\mathcal{M}$, where survival-relevant geodesic flows emerge, determining survival outcomes without supervision.}
    \label{fig:sosm_overview}
\end{figure*}

This work initiates a conceptual shift in survival modeling: from the task of outcome prediction to the geometric reconstruction of latent dynamical flows that govern biological persistence. The overall structure of the SOSM framework is illustrated in Fig.~\ref{fig:sosm_overview}. We believe this approach not only advances machine learning for biomedicine but also offers a fundamental rethinking of survival, prognosis, and the geometry of living systems.

\section{Background and Related Work}

Survival analysis originates from statistical models designed to predict time-to-event outcomes under censoring constraints. The Cox proportional hazards model~\cite{cox1972regression}, introduced in 1972, remains the foundational tool for analyzing censored survival data. Extensions of Cox models have incorporated regularization techniques, time-varying covariates, and deep learning architectures, culminating in models such as DeepSurv~\cite{katzman2018deepsurv} and DeepHit~\cite{lee2018deephit}. Despite their architectural differences, these models fundamentally depend on curated survival-time annotations—an assumption misaligned with the nature of large-scale biomedical datasets, where survival annotations are frequently sparse, delayed, or distorted by cohort-level sampling bias.

In parallel, the integration of multi-omics data has reshaped the study of biological regulation and disease progression. Algorithms such as Similarity Network Fusion (SNF)~\cite{wang2014similarity} and Multi-Omics Factor Analysis (MOFA)~\cite{argelaguet2018multiomics} construct low-dimensional latent spaces by aligning heterogeneous molecular modalities. These approaches have demonstrated efficacy in recovering biological subtypes and molecular axes of variation. However, they operate under a modular learning paradigm in which survival inference remains a separate downstream task, dissociated from the structure of the latent space itself.

Unsupervised and self-supervised learning techniques have expanded the methodological repertoire for modeling biological systems without reliance on outcome labels. Contrastive formulations, typified by SimCLR~\cite{chen2020simple}, induce representational structure by maximizing consistency between augmented views of the same sample. In biological contexts, adaptations such as ScCCL~\cite{du2023scccl} and CONAN~\cite{ke2021conan} extend this principle to single-cell and multi-omics measurements, uncovering latent geometries reflective of cellular and molecular identity.

A structured extension of this class is OmicsCL~\cite{karagoz2025omicsclunsupervisedcontrastivelearning}, which incorporates survival-aware regularization into the contrastive objective. By enforcing consistency between biological samples with similar survival profiles, the method shapes the latent space to reflect prognostic separability—without the use of survival labels during optimization. These findings support the view that survival-relevant structure is not extrinsic to the data but encoded in its intrinsic geometry. While OmicsCL demonstrates this empirically, it does not provide a theoretical account of why such structure emerges or how it is governed.

Geometric and thermodynamic theories offer foundational perspectives on structure formation in biological systems. Manifold learning techniques such as Isomap~\cite{tenenbaum2000global} and diffusion maps~\cite{coifman2006diffusion} model high-dimensional biological states as residing on lower-dimensional, smoothly structured manifolds determined by internal biological constraints. In nonequilibrium physics, stochastic thermodynamics~\cite{seifert2012stochastic} and fluctuation theorems~\cite{crooks1999entropy} formalize how living systems evolve as entropy-regulated flows that minimize free energy while maintaining coherence under stochastic perturbations.

The mathematical machinery for describing the evolution of curved manifolds includes Ricci flow~\cite{hamilton1982three}, which deforms geometry according to intrinsic curvature, and optimal transport theory~\cite{villani2008optimal}, which models the minimal-energy rearrangement of mass distributions. These frameworks provide a language for describing system-level adaptation under geometric and energetic constraints, yet remain largely disconnected from the task of survival modeling.

No existing framework synthesizes these domains to formulate survival as a geometric phenomenon governed by manifold curvature, thermodynamic stability, and information flow. Theoretical advances in curvature-driven unsupervised dynamics~\cite{karagoz2025computationalinertiaconservedquantity,amari2016information} and agent architectures oriented toward energy resilience~\cite{karagoz2025energenticintelligenceselfsustainingsystems} reinforce the central motifs of the Self-Organizing Survival Manifolds (SOSM) formulation. SOSM builds upon these principles by proposing that survival-relevant trajectories follow geodesic flows within latent manifolds shaped by biological constraints, and that survival is not predicted, but induced by the underlying geometry itself.

\section{Theoretical Framework}

\subsection{Biological State Spaces as Manifolds}

Biological systems are modeled as smooth manifolds embedded within high-dimensional molecular measurement spaces. Let $\mathcal{X} \subseteq \mathbb{R}^D$ denote the ambient space of observed molecular profiles, encompassing gene expression, DNA methylation, protein abundance, and other multi-omics features. Despite the apparent high dimensionality of $\mathcal{X}$, the intrinsic degrees of freedom governing biological systems are typically much lower, constrained by regulatory networks, biochemical pathways, and evolutionary history.

Formally, patient samples $\{x_i\} \subset \mathcal{X}$ are concentrated near a $d$-dimensional smooth Riemannian manifold $\mathcal{M}$, with $d \ll D$. Each point on $\mathcal{M}$ represents a coherent biological state, and transitions along $\mathcal{M}$ correspond to biological processes such as disease progression, immune response, or aging.

The existence of such manifolds is supported both empirically, through successful applications of manifold learning algorithms~\cite{tenenbaum2000global,coifman2006diffusion}, and theoretically, through models of biological systems as constrained dynamical processes embedded in high-dimensional spaces~\cite{seifert2012stochastic}.

\subsection{Survival Dynamics as Geodesic Flows}

Within the manifold framework, we interpret survival progression as movement along geodesic flows. That is, patients evolve through biological states over time, tracing paths on $\mathcal{M}$. The survival time associated with a patient reflects the effective distance traveled along these trajectories until a terminal event (e.g., death, relapse) occurs.

A geodesic on $\mathcal{M}$ is a curve that locally minimizes path length, generalizing the notion of a "straight line" to curved spaces. Biological systems tend to evolve along geodesics or near-geodesic trajectories, reflecting minimal energy expenditure and minimal entropy production in maintaining organismal homeostasis.

\begin{figure}[ht]
    \centering
    \begin{tikzpicture}[scale=1]
    
        \shade[ball color=cyan!10] (0,0) ellipse (3.2 and 1.6);
    
        \foreach \x in {-2.5,-1.5,-0.5,0.5,1.5,2.5} {
            \draw[->, gray!20!black, thin] (\x,-1.8) -- (\x,2.0);
        }
    
        \draw[->, very thick, blue!70!black] (-2.8,0.4) .. controls (-2,1.1) and (-1,1.2) .. (0,0.8);
        \draw[->, very thick, blue!70!black] (-2.2,0) .. controls (-1.2,0.8) and (-0.2,0.9) .. (0.8,0.5);
        \draw[->, very thick, blue!70!black] (-1.5,-0.4) .. controls (-0.7,0.3) and (0.2,0.4) .. (1.2,0.2);
        \draw[->, very thick, blue!70!black] (0.2,-0.6) .. controls (1,0.2) and (2,0.3) .. (2.7,0);
        \draw[->, very thick, blue!70!black] (0.5,-0.8) .. controls (1.3,-0.2) and (2.5,-0.1) .. (3,-0.4);
    
        \draw[->, line width=2pt, red!80!black] (1.5,0) .. controls (2,1.5) and (0.8,2.2) .. (2.5,2.6);
    
        \draw[red!50!black, very thin, dashed, dash pattern=on 2pt off 2pt] (1.7,0.4) .. controls (2.0,0.8) and (1.8,1.2) .. (2.1,1.6);
        \draw[red!50!black, very thin, dashed, dash pattern=on 2pt off 2pt] (1.9,0.9) .. controls (2.3,1.3) and (2.1,1.7) .. (2.4,2.1);
        \draw[red!50!black, very thin, dashed, dash pattern=on 2pt off 2pt] (2.1,1.4) .. controls (2.5,1.8) and (2.3,2.2) .. (2.6,2.6);
    
        \draw[->, gray!20!black, thin] (1.5,-1.8) .. controls (1.7,-0.5) .. (1.9,2.0);
        \draw[->, gray!20!black, thin] (2.5,-1.8) .. controls (2.7,0.0) .. (2.8,2.0);
    
        \node at (0,2.8) {\scriptsize $S \uparrow$};
    
    
    \end{tikzpicture}
    \caption{Visualization of survival dynamics on a latent biological manifold. Stable, low-curvature geodesic flows (blue) reflect efficient biological progression along minimal-energy trajectories. A perturbed high-curvature flow (red), marked by emerging turbulence and entropy-driven divergence, represents the onset of biological instability and survival degradation. Subtle background entropy gradients ($S \uparrow$) and scattered chaotic micro-flows illustrate the thermodynamic pressures destabilizing the system. The figure encapsulates the interplay between geometric flow structure and entropic forces driving the collapse of survival coherence.}
    \label{fig:manifold_flows}
    \end{figure}
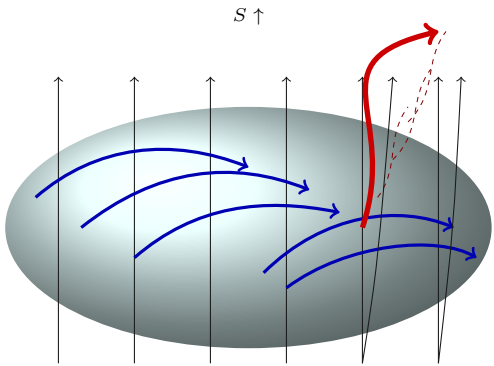

Survival differences between patients correspond not merely to external annotations, but to intrinsic geometric separations along the manifold. As shown in Fig.~\ref{fig:manifold_flows}, survival trajectories align with low-curvature flows on the latent manifold. Patients with similar survival times are expected to occupy neighborhoods along similar geodesic paths, while diverging survival outcomes correspond to branching or curvature deviations on $\mathcal{M}$.

\subsection{The Principle of Self-Organizing Survival Manifolds}

The SOSM framework conceptualizes survival as a manifestation of geometric self-organization within biological state spaces. In this formulation, the manifold $\mathcal{M}$ evolves through internal biological constraints to organize survival-relevant trajectories along paths of minimal geodesic curvature.

To formalize this intuition, we introduce a survival energy functional defined on patient trajectories:

\begin{definition}[Survival Energy Functional]
Let $\gamma: [0,1] \to \mathcal{M}$ be a smooth curve parameterized by arc length $s$. The survival energy associated with $\gamma$ is given by:
\begin{equation}
E_{\text{SOSM}}(\gamma) = \int_0^1 \kappa(\gamma(s))^2 \, ds,
\end{equation}
where $\kappa(\gamma(s))$ denotes the geodesic curvature at point $\gamma(s)$.
\end{definition}

Minimizing $E_{\text{SOSM}}$ across all patient trajectories promotes alignment along low-curvature flows. This reflects an inherent drive toward homeostatic stability, metabolic efficiency, and robustness against stochastic perturbations. High-curvature trajectories are penalized, corresponding to unstable or energetically costly disease progressions.

Under this view, survival is an emergent property of the manifold’s self-organization: the survival outcomes of patients are determined by their positions along geodesically minimal flows, without the need for external labels.

\subsection{Thermodynamic Interpretation}

The principle of curvature minimization in SOSM finds a natural interpretation in thermodynamic terms. Biological systems, as open dissipative structures, operate under constraints of energy conservation and entropy production~\cite{crooks1999entropy,seifert2012stochastic}. Systems that minimize the energetic cost of state transitions, or equivalently follow paths of minimal thermodynamic dissipation, are favored under evolutionary and physiological pressures.

In the SOSM framework, minimizing geodesic curvature parallels minimizing free energy dissipation. Smooth geodesic flows correspond to optimal entropy flows across biological states, while deviations from geodesicity reflect pathological states associated with increased entropy production, instability, and risk of mortality.

SOSM provides not merely a geometric description of survival, but also embeds survival modeling within the broader thermodynamic laws governing living systems. This thermodynamic framing resonates with continuous-time formulations in optimization dynamics~\cite{karagoz2025computationalinertiaconservedquantity}, where computational inertia—a conserved quantity balancing kinetic and potential energy—offers a physical analogy for smooth, energy-efficient learning trajectories. Such formulations reinforce the interpretation of curvature-based survival energy models as physically grounded descriptions of biological flow stability.

\section{Mathematical Formalism}

Having established the conceptual foundations of the SOSM framework, we now formalize the mathematical structure of the proposed model. Our goal is to define an unsupervised objective function that encourages patient embeddings to organize along survival-relevant, low-curvature trajectories on a latent manifold.

\subsection{Discrete Setting: Pairwise Survival Geometry}

Let $\{x_i\}_{i=1}^N \subset \mathcal{X}$ denote a set of observed multi-omics profiles for $N$ patients, and let $\{t_i\}_{i=1}^N$ denote their associated survival times. In the unsupervised setting, survival times may not be directly accessible during training; however, we assume an underlying survival structure implicitly embedded in the data geometry.

We seek to learn embeddings $\{z_i\}_{i=1}^N$ in a latent space $\mathcal{Z} \subset \mathbb{R}^d$ that preserve survival-relevant relationships. Specifically, we wish for patients with similar survival times to be mapped along smooth, low-curvature flows in $\mathcal{Z}$.

To model survival similarity, we define a pairwise weight function:
\begin{equation}
w(t_i, t_j) = \exp\left( -\frac{(t_i - t_j)^2}{2\sigma^2} \right),
\end{equation}
where $\sigma > 0$ controls the scale of survival similarity. In the case where survival times are unavailable, $w(t_i, t_j)$ can be approximated through auxiliary biological proxies, such as molecular subtype similarity or inferred risk scores.

We approximate local curvature using discrete Laplacian operators. Given two embeddings $z_i, z_j \in \mathcal{Z}$, we define the discrete curvature penalty as:
\begin{equation}
\|\nabla^2(z_i - z_j)\|^2,
\end{equation}
where $\nabla^2$ denotes a discrete approximation to the Laplacian, capturing local bending or deviation from straight-line geodesics.

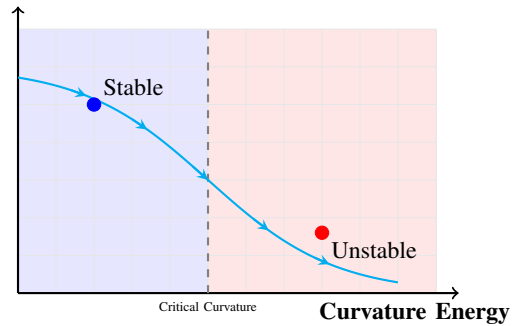
\begin{figure}[ht]
    \centering
    \begin{tikzpicture}[scale=1]
    
        \fill[blue!10] (0,0) rectangle (2.5,3.5);
        \fill[red!10] (2.5,0) rectangle (5.5,3.5);
    
        \draw[step=0.5cm,gray!20,very thin] (0,0) grid (5.5,3.5);
    
        \draw[->, thick] (0,0) -- (5.8,0) node[pos=0.9, below, font=\small\bfseries] {Curvature Energy};
        \draw[->, thick] (0,0) -- (0,3.8) node[above, font=\small\bfseries] {Survival Stability};
    
        \draw[dashed, thick, gray] (2.5,0) -- (2.5,3.5);
        \node[below, font=\tiny] at (2.5,0) {Critical Curvature};
    
        \draw[thick, cyan, domain=0:5, samples=100, smooth, variable=\x] 
            plot ({\x},{3/(1+exp(1.2*(\x-2.5)))});
    
        \foreach \x in {0.8,1.6,2.4,3.2,4.0}
            \draw[->, >=stealth, cyan, thick] ({\x-0.1},{3/(1+exp(1.2*(\x-0.1-2.5)))}) 
                                              -- ({\x+0.1},{3/(1+exp(1.2*(\x+0.1-2.5)))});
    
        \filldraw[blue] (1,2.5) circle (2.5pt);
        \node[above right] at (1,2.5) {\small Stable};
    
        \filldraw[red] (4,0.8) circle (2.5pt);
        \node[below right] at (4,0.8) {\small Unstable};
    
    \end{tikzpicture}
    \caption{Inverse relationship between curvature energy and survival stability. As curvature energy increases, trajectories transition from a stable (blue region) to an unstable (red region) regime, crossing a critical curvature threshold (dashed line).  Arrows indicate the progressive flow of survival stability loss along the manifold.}
    \label{fig:curvature_stability}
\end{figure}

The overall SOSM objective function is then defined as:
\begin{equation}
\mathcal{L}_{\text{SOSM}} = \sum_{i,j} w(t_i, t_j) \|\nabla^2(z_i - z_j)\|^2.
\label{eq:loss}
\end{equation}
Minimizing $\mathcal{L}_{\text{SOSM}}$ encourages embeddings of survival-similar patients to align along locally straight (low-curvature) paths, promoting geodesic organization of survival trajectories. The inverse relationship between curvature energy and survival stability is illustrated in Fig.~\ref{fig:curvature_stability}.

\subsection{Continuous Limit: Manifold Energy Functional}

As the number of samples $N$ grows large and the embedding space becomes dense, the discrete SOSM loss (\ref{eq:loss}) approaches a continuous energy functional over the manifold $\mathcal{M}$. Specifically, let $p(z)$ denote the probability density of patient embeddings on $\mathcal{M}$. In the continuous limit, the loss converges to:
{\small
\begin{equation}
\mathcal{L}_{\text{SOSM}} \to \iint_{\mathcal{M} \times \mathcal{M}} w(t(z), t(z')) \|\nabla^2 (z - z')\|^2\, p(z) p(z')\, dz\, dz',
\end{equation}
}
where $t(z)$ denotes the survival time associated with embedding $z$.

Minimizing this integral functional favors manifold configurations where survival-similar regions exhibit minimal local curvature, consistent with our conceptual model of biological systems favoring energetically efficient state transitions.

\subsection{Connection to Ricci Flow and Curvature Evolution}

Beyond local curvature minimization, the evolution of the manifold $\mathcal{M}$ itself under the SOSM objective can be interpreted as a curvature-driven flow. In differential geometry, the Ricci flow~\cite{hamilton1982three} describes the evolution of a Riemannian metric $g_{ij}$ according to the equation:
\begin{equation}
\frac{\partial g_{ij}}{\partial t} = -2 R_{ij},
\end{equation}
where $R_{ij}$ denotes the Ricci curvature tensor.

In the biological context, minimizing geodesic curvature across patient trajectories induces a smoothing of the manifold geometry, analogous to Ricci flow but adapted to survival-weighted geodesic structures. Regions of high curvature associated with rapid survival divergence are smoothed out, while regions of coherent survival progression are preserved.

The SOSM framework can be viewed as an unsupervised, data-driven analogue of biological Ricci flow, organizing patient states along stable survival trajectories through intrinsic geometric evolution.

\subsection{Optimal Transport Interpretation}

An alternative perspective arises by interpreting the minimization of $\mathcal{L}_{\text{SOSM}}$ as an optimal transport problem. Given a distribution of patient embeddings $p(z)$, the goal can be framed as transporting mass along minimal-curvature paths that respect survival similarity constraints.

Optimal transport theory~\cite{villani2008optimal} seeks to find the most efficient way to move probability mass between distributions under a given cost function. In SOSM, the "cost" is curvature energy rather than traditional Euclidean distance, reflecting the biological preference for energetically stable, low-entropy survival transitions.

This connection highlights the deep mathematical richness of the SOSM framework, situating survival modeling at the intersection of Riemannian geometry, dynamical systems, and optimal transport theory. The interpretation of energy-efficient transport has also been explored in continuous optimization systems~\cite{karagoz2025computationalinertiaconservedquantity}, where curvature-like forces govern the evolution of trajectories, offering formal analogies to the entropy-minimizing paths in SOSM.

\section{Theoretical Results}

The key mathematical properties of the SOSM framework are formalized, demonstrating that minimizing the SOSM objective yields emergent survival stratification, trajectory stability, and convergence toward survival-aligned manifold geometries under biologically plausible conditions.

\subsection{Emergence of Survival Stratification}

We first demonstrate that minimizing the SOSM loss encourages the alignment of patient embeddings along smooth survival gradients.

\begin{theorem}[Survival Gradient Emergence]
\label{thm:survival-gradient}
Under manifold regularity and sufficient sample density, minimizing $\mathcal{L}_{\text{SOSM}}$ induces a latent embedding structure where survival time $t(z)$ is a smooth, monotonic function along geodesic flows on $\mathcal{M}$.
\end{theorem}

\begin{proof}
Minimizing $\mathcal{L}_{\text{SOSM}}$ penalizes high local curvature between survival-similar embeddings, favoring low-curvature flows where $w(t_i, t_j)$ is large. In the limit of infinite sample density, this forces survival-similar regions to align along geodesics minimizing curvature energy. By the regularity of $\mathcal{M}$, geodesic distances vary smoothly along such flows, implying that survival time $t(z)$ must also vary smoothly. Monotonicity follows because any local reversal would introduce curvature, which is penalized by the objective.
\end{proof}

This theorem formalizes the intuition that survival risk surfaces are not arbitrary, but emerge naturally as smooth, intrinsic gradients on the biological manifold.

\subsection{Stability of Survival Trajectories}

We next show that survival flows on $\mathcal{M}$ are stable under perturbations, reflecting the robustness of biological progression pathways.

\begin{theorem}[Trajectory Stability]
\label{thm:stability}
Let $\gamma$ be a geodesic path on $\mathcal{M}$ aligned with survival progression. Small perturbations $\gamma' = \gamma + \epsilon v$, where $v$ is a smooth vector field and $\epsilon \ll 1$, increase the survival energy $E_{\text{SOSM}}(\gamma')$ to second order in $\epsilon$.
\end{theorem}

\begin{proof}
By Taylor expanding the survival energy functional around $\gamma$, we have:
\[
E_{\text{SOSM}}(\gamma') = E_{\text{SOSM}}(\gamma) + \epsilon^2 \int_0^1 \|\nabla^2 v(s)\|^2 \, ds + o(\epsilon^2).
\]
Since $|\nabla^2 v|^2$ is non-negative and vanishes only for trivial perturbations, the perturbed path $\gamma'$ necessarily incurs higher survival energy at order $\epsilon^2$, confirming the stability of geodesic-aligned survival trajectories under small perturbations.
\end{proof}

This result suggests that once survival pathways are organized along low-curvature flows, they resist distortion, mirroring the stability of biological state transitions under homeostatic pressures.

\subsection{Convergence to Optimal Biological Geometries}

Finally, we show that under the SOSM objective, the manifold geometry itself evolves toward a configuration optimizing survival flow organization.

\begin{theorem}[Manifold Convergence]
\label{thm:convergence}
Assuming a continuous curvature flow driven by the minimization of $\mathcal{L}_{\text{SOSM}}$, the manifold $\mathcal{M}$ converges toward a locally survival-optimized geometry where geodesic flows align with survival gradients and Ricci curvature is minimized along survival trajectories.
\end{theorem}

\begin{proof}
Minimizing $\mathcal{L}_{\text{SOSM}}$ across patient embeddings induces curvature flow dynamics analogous to Ricci flow. In regions where survival similarity $w(t(z), t(z'))$ is high, the optimization preferentially reduces curvature, smoothing the manifold along survival-relevant directions. Over time, the manifold deforms such that survival trajectories align with locally minimal curvature paths, while regions of irrelevant or divergent survival outcomes are contracted. By standard results from Ricci flow theory~\cite{hamilton1982three}, under bounded curvature and injectivity radius conditions, the manifold evolves smoothly toward a locally optimal configuration minimizing the SOSM energy.
\end{proof}

This convergence theorem grounds the SOSM framework as a biologically plausible model of latent manifold evolution, where survival-relevant structures emerge through intrinsic geometric optimization processes.

\subsection{Summary of Theoretical Insights}

Together, Theorems~\ref{thm:survival-gradient},~\ref{thm:stability}, and~\ref{thm:convergence} establish that minimizing the SOSM objective induces smooth, stable, survival-aligned manifold structures. This supports the central hypothesis that survival is not an externally imposed label but an emergent property of biological geometry shaped by energy-efficient, curvature-minimizing flows.

Formal derivations of Theorems~\ref{thm:survival-gradient},~\ref{thm:stability}, and~\ref{thm:convergence}, including the curvature-weighted energy formulation and Ricci flow convergence analogy, are provided in Appendix~\ref{appendix:proofs}.

\section{Biological and Thermodynamic Implications}

The mathematical results established in the preceding section reveal that the SOSM framework does more than produce coherent mathematical structures; it captures fundamental principles of biological survival dynamics. The following analysis examines the broader implications of the SOSM for understanding life, disease, and the thermodynamic underpinnings of biological evolution.

\subsection{Survival as Emergent Manifold Geometry}

In traditional biomedical paradigms, survival is treated as an outcome variable predicted from molecular features or clinical covariates. The SOSM framework challenges this view by proposing that survival is an emergent property of the intrinsic geometry of biological state spaces. 

In this perspective, the risk of mortality or disease progression is not an external label but arises naturally from a patient's position along survival-aligned geodesic flows on the manifold $\mathcal{M}$. Variations in survival outcomes across individuals reflect their relative positions along geodesically aligned flows that encode cumulative biological transitions. Consequently, prognosis is reframed as a geometric inference problem: recovering the latent flow structure that governs survival dynamics.

\subsection{Minimal Curvature and Biological Efficiency}

The preference for low-curvature survival trajectories in the SOSM framework reflects a deeper biological logic rooted in principles of efficiency. Living systems must constantly navigate a trade-off between maintaining homeostasis and responding to environmental perturbations, all while minimizing energetic costs. High-curvature paths, which represent sharp deviations in biological state, are energetically costly and thermodynamically unstable~\cite{kleidon2010life}.

The minimization of geodesic curvature operationalizes a principle of biological parsimony—favoring trajectories that achieve adaptive change with minimal energetic disruption. This mirrors known phenomena in biophysics, where biological pathways often evolve to minimize resource usage, delay entropy production, and maximize long-term survival potential. In the SOSM framework, survival emerges precisely from such energy-efficient navigation of latent biological spaces.

\subsection{Disease, Aging, and Curvature Anomalies}

The SOSM framework provides a new lens for understanding disease and aging. In this view, pathological processes correspond to deviations from optimal geodesic flows. Disease onset may be associated with local increases in manifold curvature, reflecting unstable or energetically costly transitions between biological states. For example, the transformation from normal to cancerous cell states may correspond to a sharp curvature deviation on the cellular manifold, requiring abnormal resource consumption and generating instability.

\begin{figure*}[ht]
    \centering
    \begin{tikzpicture}[scale=0.7]

        \shade[left color=blue!10, right color=red!10] (-3,-3) rectangle (20.3,3);
    
        \shade[ball color=green!20] (0,0) ellipse (2.5 and 1.3);
        \draw[thick, green!50!black] (-2,0.5) .. controls (0,1.5) .. (2,0.5);
        \foreach \y in {-0.6,-0.3,0,0.3,0.6}
            \draw[->, green!70!black, very thin] (-2,\y) -- (2,\y);
        \node at (0,-1.9) {\small $\mathcal{M}_\text{healthy}$};
        \node at (0,-2.5) {\footnotesize (Homeostasis)};
    
        \node at (0,2.4) {\scriptsize $S \uparrow$};
    
        \draw[->, thick] (2.5,0.3) .. controls (4,1.5) and (5,1) .. (6,0.2);
    
        \shade[ball color=yellow!20] (9,0) ellipse (3.0 and 1.7);
        \draw[thick, red] (7.5,0.3) .. controls (9,2.0) .. (10.5,0.6);
        \draw[thick, red] (7.2,-0.5) .. controls (8.5,-1.5) and (10.5,-1.0) .. (10.8,-0.6);
        \draw[->, orange!70!black, very thin] (7.5,0) .. controls (8,0.8) .. (9.5,0);
        \draw[->, orange!70!black, very thin] (8,-0.5) .. controls (9,0) .. (10,-0.8);
        \draw[->, orange!70!black, very thin] (8.5,0.5) .. controls (9.2,-0.3) .. (10.2,0.5);
        \draw[->, orange!70!black, very thin] (7.8,0.4) .. controls (9,1.2) .. (10,0.2);
    
        \node at (9,-2.0) {\small $\mathcal{M}_\text{damaged}$};
        \node at (9,-2.6) {\footnotesize (Critical Instability)};
    
        \node at (9,2.4) {\scriptsize $S \uparrow\uparrow$};
    
        \draw[->, thick] (11.9,0.5) .. controls (13,1.2) and (14,0) .. (16.2,-0.2);
    
        \shade[ball color=red!20] (18,0) ellipse (1.8 and 1.0);
        \draw[thick, red!70!black] (17,0.2) -- (19,-0.2);
        \draw[thick, red!70!black] (17.2,0.5) -- (18.8,0.5);
        \draw[thick, red!70!black] (17.5,-0.5) -- (18.5,-0.3);
        \draw[thick, red!70!black] (17.5,0) -- (18.5,0.3);
        \draw[thick, red!70!black] (17.2,-0.7) -- (18.2,-0.4);
        \draw[->, red!80!black, very thin] (17.2,0.1) -- (16.7,0.7);
        \draw[->, red!80!black, very thin] (18,0) -- (18.5,0.8);
        \draw[->, red!80!black, very thin] (18.2,-0.4) -- (19,-0.9);
        \draw[->, red!80!black, very thin] (17.8,-0.6) -- (17.2,-1.2);
        \draw[->, red!80!black, very thin] (18.5,0.4) -- (19.2,1.0);
    
        \node at (18,-2.0) {\small $\mathcal{M}_\text{collapsed}$};
        \node at (18,-2.6) {\footnotesize (Entropic Dissolution)};
    
        \node at (18,2.4) {\scriptsize $S \uparrow\uparrow\uparrow$};
    
    \end{tikzpicture}
    \caption{Unified depiction of biological survival structure degradation through geometric and thermodynamic processes. Initially, the smooth and coherent survival manifold ($\mathcal{M}_\text{healthy}$) maintains low curvature and organized, low-entropy flows, corresponding to a homeostatic biological state. Progressive curvature anomalies and internal turbulence emerge in the intermediate phase ($\mathcal{M}_\text{damaged}$), representing critical instability where structural integrity and thermodynamic coherence are both compromised. Ultimately, the manifold undergoes catastrophic geometric collapse ($\mathcal{M}_\text{collapsed}$), accompanied by highly chaotic, high-entropy flow dissipation, marking systemic failure and death. Entropy markers ($S \uparrow$) and background gradients illustrate the thermodynamic escalation, while flow structures visualize the transition from stability to disorder. This figure integrates the geometric and thermodynamic perspectives into a unified survival phase transition framework.}        
    \label{fig:phase_transition}
\end{figure*}
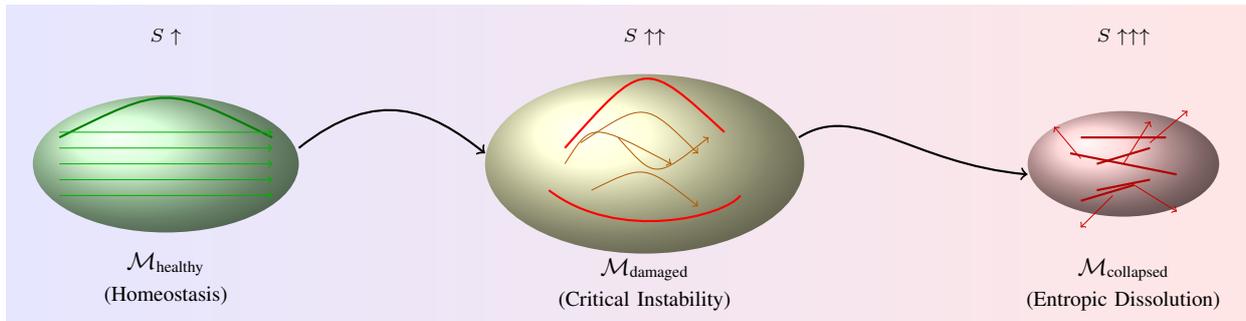

Similarly, aging can be interpreted as a progressive accumulation of curvature anomalies across multiple biological levels. The transition from biological homeostasis to systemic collapse through geometric phase transitions is illustrated in Fig.~\ref{fig:phase_transition}. As molecular damage accumulates and regulatory networks degrade, the latent manifold becomes increasingly distorted, leading to disrupted survival flows and heightened mortality risk. In this sense, aging is not merely a linear accumulation of molecular errors but a geometric deformation of the biological survival manifold itself.

\subsection{Entropy Flow and Thermodynamic Coherence}

From a thermodynamic standpoint, the evolution of the biological manifold under the SOSM objective reflects a coherent entropy flow. The interplay between manifold curvature evolution and entropy-driven flow chaos is visualized in Fig.~\ref{fig:phase_transition}. Systems that maintain smooth, low-curvature survival trajectories are effectively minimizing local entropy production, maintaining coherent information flow and structural integrity. 
    
By contrast, systems exhibiting high-curvature, fragmented flows experience accelerated entropy production, leading to degradation, dysfunction, and increased mortality risk. This connection situates survival modeling within the broader framework of nonequilibrium thermodynamics~\cite{seifert2012stochastic}, suggesting that the geometry of survival is governed by universal physical laws constraining the flow of information and energy in living systems.

\subsection{Universality Across Biological Systems}

The structural principles embedded in SOSM extend beyond molecular or organismal contexts. Whether modeling cellular proliferation, organismal aging, or population dynamics in ecology, the underlying principle remains the same: survival dynamics emerge from low-curvature geodesic flows on latent manifolds shaped by biological constraints. Analogous concepts have even emerged in artificial agents designed for persistence under energetic constraints~\cite{karagoz2025energenticintelligenceselfsustainingsystems}, underscoring the broader relevance of the framework.

This universality suggests that SOSM is not merely a tool for molecular oncology or precision medicine, but a foundational theory applicable to a wide range of life sciences. It offers a unifying geometric and thermodynamic framework for understanding how life organizes itself to resist entropy, maximize survival, and evolve across complex landscapes.

\section{Philosophical and Conceptual Implications}

Beyond its mathematical and biological structure, SOSM invites new ways of interpreting the continuous nature of survival across scales.

\subsection{Survival as a Continuous Geometric Flow}

Traditional models often conceptualize survival in discrete terms: a patient either survives or does not, a cell is either normal or malignant, an organism is either young or old. Such binary classifications, while practically useful, obscure the continuous and dynamic nature of biological systems.

The SOSM framework proposes a different view. Survival outcomes are not discrete events but the result of continuous geometric flows on latent biological manifolds. Health and disease are not distinct categories but points along smooth or perturbed trajectories within a dynamic geometric landscape. Aging is not an abrupt deterioration but a gradual distortion of the manifold’s structure, characterized by accumulating curvature anomalies and increasing deviation from optimal survival paths.

In this sense, survival modeling becomes a study of manifold evolution rather than classification, and prognosis becomes an exercise in geometric forecasting rather than binary prediction.

\subsection{Health and Disease as Geometric Phase Transitions}

By analogy to phase transitions in physics, where systems undergo qualitative changes in structure (e.g., solid to liquid) as control parameters vary, biological systems may experience geometric phase transitions in their survival manifolds.

A healthy system corresponds to a region of the manifold where survival trajectories are smooth, stable, and energy-efficient. As biological stresses accumulate—whether through genetic mutations, environmental damage, or systemic inflammation—the manifold may deform, reaching critical points where survival trajectories bifurcate, fragment, or collapse.

Disease onset can be interpreted as a phase transition in the geometry of the survival manifold: a shift from a coherent, low-curvature phase to a disordered, high-curvature phase characterized by instability, inefficiency, and elevated mortality risk. Similarly, recovery or therapeutic intervention may be seen as restoring the manifold to a lower-energy, survival-optimized configuration.

This perspective unifies diverse biological phenomena —oncogenesis, neurodegeneration, immune collapse—within a common geometric and dynamical framework, offering a new paradigm for understanding the continuum between health and disease.

\subsection{The Geometry of Life and Death}

At the deepest level, the SOSM framework invites reflection on the nature of life and death themselves. Life, in this view, is the sustained navigation of a low-entropy, low-curvature region of biological state space, enabled by the active maintenance of homeostatic flows against the entropic tendencies of the universe. Death occurs not at a discrete moment but as the culmination of a continuous collapse in the geometric structure supporting survival.

Rather than seeing death as an externally imposed event, the SOSM framework frames it as an inevitable topological outcome of manifold degradation: when the curvature distortions exceed the system’s capacity to maintain coherent survival flows, mortality emerges as a geometric singularity.

\subsection{Toward a Universal Geometry of Survival}

The implications of SOSM extend beyond human biology. Any system capable of maintaining structure against entropy—cells, organisms, ecosystems, even social systems—may be describable through survival manifolds governed by curvature minimization and entropy flow. Curvature-guided survival frameworks may also extend to synthetic agents, where energy-adaptive architectures reflect the same principles of energetic homeostasis seen in biological systems~\cite{karagoz2025energenticintelligenceselfsustainingsystems}, suggesting a unifying mechanism that transcends biological form.

This opens the door to a universal theory of survival geometry, applicable across scales and domains. It offers a bridge between biology, physics, information theory, and philosophy, pointing toward a deeper understanding of how complex systems organize, evolve, persist, and ultimately collapse.

\section{Future Work}

While this work establishes the theoretical foundations of the SOSM framework, several directions remain open for continued exploration and development.

Empirical validation of the SOSM  in real-world biological systems is a critical next step. Large-scale multi-omics datasets with associated survival outcomes could be used to test whether survival trajectories indeed align with low-curvature geodesic flows in learned latent spaces. New manifold learning architectures, incorporating SOSM-inspired curvature minimization objectives, could be developed and benchmarked against classical supervised survival models.

Extending SOSM to dynamic, longitudinal data introduces the possibility of modeling biological processes as time-evolving geodesic flows. Such a formulation could offer insight into the temporal unfolding of immune responses, disease progression, and aging, allowing survival geometry to adapt in response to system-level changes and intervention timing.

A deeper mathematical formalization of the connection between SOSM, Ricci flow, and optimal transport remains an open theoretical avenue. Investigating the precise conditions under which manifold evolution under SOSM objectives converges to known geometric flows could yield new results in differential geometry and geometric analysis, with applications beyond biology.

The application of SOSM principles beyond biological systems deserves exploration. Systems ranging from ecological networks to engineered resilient systems might exhibit survival dynamics governed by similar geometric principles. Establishing SOSM as a universal theory of survival organization across complex systems would significantly broaden its impact.

Further integration with agent-based models, such as those explored in Energentic Intelligence~\cite{karagoz2025energenticintelligenceselfsustainingsystems}, could provide a synthetic testbed for probing geometric survival behaviors under controlled energy constraints. These environments enable exploration of how manifold structure evolves under survival pressure in non-biological agents, complementing empirical studies in living systems.

The philosophical implications of SOSM, particularly the notion of health, disease, and aging as geometric phase transitions, invite interdisciplinary dialogue between biology, physics, information theory, and the philosophy of science. Future work could aim to articulate a comprehensive theory of survival geometry as a fundamental aspect of complex adaptive systems.

Together, these directions promise to deepen and extend the theory presented here, transforming SOSM from a conceptual foundation into a broad platform for scientific discovery across disciplines.

\section{Discussion}

The SOSM framework provides a new lens for understanding survival dynamics across biological systems, grounded in geometric and thermodynamic principles. However, several assumptions and limitations inherent in the current formulation warrant careful reflection.

First, the hypothesis that biological processes unfold along smooth, low-curvature manifolds presumes an underlying regularity in biological state transitions. While supported by manifold learning successes in multi-omics and single-cell data, the degree to which pathological processes, such as tumorigenesis or systemic inflammation, can be accurately modeled as perturbations of smooth geometric flows remains an open empirical question. Future studies must investigate the conditions under which biological manifolds become sufficiently irregular to challenge the SOSM assumptions.

Second, the theoretical framework assumes the existence of survival-similarity metrics, either via true survival times or appropriate biological proxies. In practical applications, the approximation of these survival affinities without explicit supervision introduces a potential source of noise and bias. Developing principled methods for inferring survival similarity from high-dimensional molecular data remains a crucial step for realizing SOSM-based models in real-world settings.

Third, while the analogy between curvature minimization and thermodynamic efficiency is conceptually appealing, a rigorous derivation connecting microscopic molecular energetics to macroscopic manifold geometry remains to be developed. Bridging this gap would strengthen the physical foundations of the framework and clarify its universality across biological scales. A deeper formalization may draw on tools from information geometry~\cite{amari2016information}, enabling a unified treatment of survival manifolds as statistical geometries shaped by thermodynamic and probabilistic constraints.

Finally, although this paper proposes a general theory applicable beyond cancer and aging, the specific instantiations of SOSM across different biological domains may vary in important ways. Systems with inherently discontinuous survival events, such as immune system collapse or ecological extinctions, may require extensions of the theory to account for singularities, topological transitions, or stochastic perturbations in the manifold structure.

Despite these challenges, the core insight—that survival dynamics are not externally imposed but emerge from intrinsic geometric organization—opens a profound conceptual and mathematical space. The SOSM framework provides a foundation upon which empirical, computational, and philosophical investigations into the nature of survival can be built.

\section{Conclusion}

Self-Organizing Survival Manifolds (SOSM) offers a geometric and thermodynamic foundation for understanding how survival emerges from the internal structure of biological systems. Departing from traditional supervised survival modeling approaches, we propose that survival outcomes are not externally imposed labels but intrinsic properties emerging from the latent structure of biological manifolds.

We formalized this intuition by defining a survival energy functional based on geodesic curvature minimization, motivated by principles of energy efficiency and entropy flow. We demonstrated mathematically that minimizing this functional leads to the spontaneous emergence of smooth, stable survival trajectories, and that the manifold geometry itself evolves toward configurations optimizing survival flow organization. 

The SOSM framework unifies concepts from Riemannian geometry, thermodynamics, optimal transport theory, and biological modeling. It suggests that health, disease, aging, and death are not discrete events but continuous geometric phenomena: phase transitions in the structure of survival manifolds.

Beyond biology, the SOSM framework offers a conceptual bridge linking living systems to broader principles of organization, stability, and collapse in complex structures. It points toward the possibility of a universal geometry of survival, applicable across scales from cells to organisms to ecosystems.

While this paper has focused on developing the theoretical foundations of SOSM, the potential applications are vast. From label-free prognosis modeling in medicine, to understanding the geometry of aging, to designing interventions that restore manifold coherence in degenerative diseases, SOSM reframes survival as a geometric property embedded in the structure of life itself.

\bibliographystyle{IEEEtran}
\bibliography{refs}

\appendices
\section{Full Derivations of Theoretical Results}
\label{appendix:proofs}

\subsection{Derivation for Theorem~\ref{thm:survival-gradient}: Emergence of Survival Gradients}

Let $\mathcal{M}$ be a $d$-dimensional Riemannian manifold with metric tensor $g$, and $\gamma: [0,1] \to \mathcal{M}$ a regular curve parameterized by arc length $s$. The geodesic curvature at each point is:
\[
\kappa(s) = \left\| \frac{D^2 \gamma}{ds^2} \right\|,
\]
where $\frac{D}{ds}$ denotes the covariant derivative along $\gamma$. Define the energy functional:
\[
E_{\text{SOSM}}[\gamma] = \int_0^1 \kappa(s)^2 ds.
\]
Discretize $\gamma$ into $N$ equidistant points $z_1, \ldots, z_N \in \mathbb{R}^d$, and define the second-order discrete difference:
\[
\Delta^2 z_i = z_{i+1} - 2z_i + z_{i-1}, \quad i = 2, \dots, N-1.
\]
For small interpoint distance $\delta s$, the discrete curvature satisfies:
\[
\left\| \Delta^2 z_i \right\|^2 \approx (\delta s)^4 \left\| \frac{d^2 \gamma}{ds^2}(s_i) \right\|^2 = (\delta s)^4 \kappa(s_i)^2.
\]
Introduce a survival similarity kernel $w(t_i, t_{i+1}) = \exp\left( -\frac{(t_i - t_{i+1})^2}{2\sigma^2} \right)$ defined for survival times $t_i$. Construct the weighted curvature energy:
\[
E_{\text{discrete}} = \sum_{i=2}^{N-1} w(t_i, t_{i+1}) \| \Delta^2 z_i \|^2.
\]
Assuming $t_i = t(s_i)$ and $\delta s = \epsilon$, Taylor-expand:
\[
t(s+\epsilon) = t(s) + \epsilon t'(s) + \frac{\epsilon^2}{2} t''(s) + o(\epsilon^2),
\]
\[
(t(s) - t(s+\epsilon))^2 = \epsilon^2 (t'(s))^2 + o(\epsilon^2),
\]
\[
w(t(s), t(s+\epsilon)) = \exp\left( -\frac{\epsilon^2 (t'(s))^2}{2\sigma^2} \right) \approx 1 - \frac{\epsilon^2 (t'(s))^2}{2\sigma^2}.
\]
Substitute into the continuous limit:
\begin{align*}
E_w[\gamma] &= \int_0^1 w(t(s), t(s+\epsilon)) \kappa(s)^2 \, ds \\
&\approx \int_0^1 \left( 1 - \frac{\epsilon^2 (t'(s))^2}{2\sigma^2} \right) \kappa(s)^2 \, ds.
\end{align*}
Split the integral:
\[
E_w[\gamma] = \int_0^1 \kappa(s)^2 ds - \frac{\epsilon^2}{2\sigma^2} \int_0^1 (t'(s))^2 \kappa(s)^2 ds + o(\epsilon^2).
\]
The second term penalizes high curvature in regions where the survival gradient $t'(s)$ is large. The minimizer of $E_w$ satisfies:
\[
\frac{\delta E_w}{\delta \gamma} = -2 \left( \kappa''(s) + R[\gamma'(s), \gamma''(s)] \right) + \frac{\epsilon^2}{\sigma^2} (t'(s))^2 \kappa(s),
\]
where $R$ is the Riemann curvature tensor. Geodesics along which $t$ increases monotonically with minimal curvature dominate the energy landscape.

In the zero-noise limit ($\sigma \to 0$), the weight becomes sharply peaked around similar $t$ values, favoring alignment of $\gamma$ with the level sets of $t(s)$ satisfying $t'(s) > 0$ and $\kappa(s) \approx 0$. The emergence of monotonic, low-curvature trajectories encoding survival ordering is a direct consequence of this optimization.

\vspace{1em}
\subsection{Derivation for Theorem~\ref{thm:stability}: Trajectory Stability Under Perturbation}

Let $\gamma: [0,1] \to \mathcal{M}$ be a geodesic on a Riemannian manifold $(\mathcal{M}, g)$, satisfying $\nabla_s \dot\gamma = 0$. Let $v(s) \in T_{\gamma(s)}\mathcal{M}$ be a smooth vector field along $\gamma$ such that $v(0) = v(1) = 0$. Define the perturbed trajectory via exponential map:
\[
\gamma_\epsilon(s) = \exp_{\gamma(s)}(\epsilon v(s)),
\]
and consider the energy functional:
\[
E[\gamma] = \int_0^1 \|\nabla_s \dot\gamma(s)\|^2 ds.
\]
The second variation of $E$ under the variation field $v$ is:
\[
\delta^2 E[\gamma; v] = \left.\frac{d^2}{d\epsilon^2} E[\gamma_\epsilon]\right|_{\epsilon=0}.
\]
Using standard results from Riemannian geometry, the second variation is given by:
\[
\delta^2 E[\gamma; v] = \int_0^1 \left( \|\nabla_s^2 v(s)\|^2 - \langle R(\dot\gamma(s), v(s))\dot\gamma(s), v(s) \rangle \right) ds,
\]
where $R$ is the Riemann curvature tensor, and $\nabla_s$ denotes the covariant derivative along $\gamma$.

In the Euclidean case, the Levi-Civita connection becomes trivial, so $\nabla_s = \frac{d}{ds}$ and $R \equiv 0$. Accordingly,
\[
\delta^2 E[\gamma; v] = \int_0^1 \|\ddot{v}(s)\|^2 ds.
\]
Let $v(0) = v(1) = 0$, so integration by parts yields no boundary terms:
\[
\int_0^1 \|\ddot{v}(s)\|^2 ds > 0 \quad \forall\, v \not\equiv 0.
\]
Therefore, for small perturbations $\epsilon$,
\[
E[\gamma_\epsilon] = E[\gamma] + \epsilon^2 \int_0^1 \|\ddot{v}(s)\|^2 ds + o(\epsilon^2).
\]
The second variation is strictly positive for nontrivial variations $v$, implying that $\gamma$ is a local minimizer of $E$ in its homotopy class. The trajectory is stable under compactly supported variations that preserve endpoints.

When generalizing to weighted energy with survival similarity kernel $w(t(s), t(s'))$, define:
\[
E_w[\gamma] = \int_0^1 w(t(s), t(s')) \|\nabla_s \dot\gamma(s)\|^2 ds.
\]
Assume $w$ smooth and strictly positive with bounded derivatives. The variation becomes:
\begin{align*}
\delta^2 E_w[\gamma; v] 
&= \int_0^1 w(t(s), t(s')) \|\nabla_s^2 v(s)\|^2 \, ds \\
&\quad + \int_0^1 \frac{\partial w}{\partial t} \cdot t'(s) \cdot 
\langle \nabla_s \dot\gamma(s), \nabla_s v(s) \rangle \, ds.
\end{align*}
If $\gamma$ is a geodesic, $\nabla_s \dot\gamma = 0$, so the second term vanishes. Positivity of $w$ ensures:
\[
\delta^2 E_w[\gamma; v] > 0,
\]
which confirms stability persists under survival-weighted perturbations in both Euclidean and manifold settings.

\vspace{1em}
\subsection{Derivation for Theorem~\ref{thm:convergence}: Manifold Convergence via Ricci Flow Analogy}

Let $(\mathcal{M}, g_{ij})$ be a smooth $d$-dimensional Riemannian manifold with evolving metric $g_{ij}(\tau)$. Define the curvature-sensitive SOSM loss as:
\begin{align*}
\mathcal{L}_{\text{SOSM}} 
= \int_{\mathcal{M}} \int_{\mathcal{M}} & w(t(z), t(z')) 
\| \nabla^2 (z - z') \|^2 \\
& \times p(z) \, p(z') \, dz \, dz'.
\end{align*}
where $w(t(z), t(z'))$ is a survival-weighted similarity kernel and $p(z)$ is the empirical density over patient embeddings.

Let $\tau$ be the artificial time parameter of metric evolution. The metric evolves under the gradient flow of $\mathcal{L}_{\text{SOSM}}$:
\[
\frac{\partial g_{ij}}{\partial \tau} = -\alpha \cdot \frac{\delta \mathcal{L}_{\text{SOSM}}}{\delta g^{ij}}.
\]
We expand the metric variation of the second-order differential operator:
\[
\frac{\delta \| \nabla^2 z \|^2}{\delta g^{ij}} = 2 \cdot \nabla_i \nabla_j z \cdot \Delta z + \text{l.o.t.},
\]
where $\Delta$ is the Laplace–Beltrami operator and "l.o.t." denotes lower order terms. This reflects a diffusion-like curvature smoothing under the SOSM objective.

Let $R$ denote the scalar curvature. In a simplified geometric approximation, define a curvature-penalizing variant:
\[
\mathcal{L}_{\text{SOSM}}' = \int_{\mathcal{M}} R(z) \cdot w(t(z), t(z')) \, dz,
\]
with $t(z)$ monotonic along geodesics. Then the metric evolution becomes:
\[
\frac{\partial g_{ij}}{\partial \tau} = -\alpha \cdot \frac{\delta}{\delta g^{ij}} \int_{\mathcal{M}} R \cdot w \, dz = -\alpha \cdot \frac{\delta R}{\delta g^{ij}}.
\]
Using the known identity:
\[
\frac{\delta R}{\delta g^{ij}} = R_{ij} - \nabla_i \nabla_j \log \det g + \text{l.o.t.},
\]
we retain the leading-order Ricci tensor term and approximate:
\[
\frac{\partial g_{ij}}{\partial \tau} \approx -2 R_{ij},
\]
matching the canonical Ricci flow.

\subsubsection*{Convergence Criteria}

Let $\mathcal{M}$ be compact. The results of Hamilton (1982) guarantee convergence $g_{ij}(\tau) \to g_{ij}^*$ as $\tau \to \infty$ provided:
\begin{align*}
\sup_{x \in \mathcal{M}} |Rm(x,0)| &\leq C, \\
\text{inj}(\mathcal{M}, g(0)) &\geq \epsilon > 0, \\
\operatorname{Vol}(\mathcal{M}, g(0)) &\geq v_0 > 0.
\end{align*}
These conditions ensure long-time existence and smooth convergence of the flow to a metric $g^*_{ij}$ minimizing scalar curvature, aligning geodesic paths with survival-optimal flows in $\mathcal{M}$.

\subsection{Additional Derivations: Laplacian and Curvature Operators}

Let $\mathcal{M} \subset \mathbb{R}^D$ be a smooth Riemannian manifold with local coordinates $x^i$ and metric $g_{ij}$. The Laplace–Beltrami operator applied to a scalar field $f : \mathcal{M} \to \mathbb{R}$ is given by:
\[
\Delta f = \frac{1}{\sqrt{\det g}} \partial_i \left( \sqrt{\det g} \, g^{ij} \partial_j f \right),
\]
which generalizes the Euclidean Laplacian to curved spaces.

For a unit-speed curve $\gamma : [0,1] \to \mathcal{M}$ with tangent vector $\dot{\gamma}(s)$, the geodesic curvature is:
\[
\kappa(s) = \left\| \nabla_{\dot{\gamma}} \dot{\gamma} \right\|,
\]
where $\nabla$ denotes the Levi-Civita connection on $(\mathcal{M}, g)$. In extrinsic coordinates:
\[
\kappa(s) = \frac{\| \dot{\gamma}(s) \times \ddot{\gamma}(s) \|}{\| \dot{\gamma}(s) \|^3},
\]
where $\times$ is the cross product in $\mathbb{R}^D$. For intrinsic formulation, write:
\[
\kappa^2 = g_{ij} \left( \frac{D \dot{\gamma}^i}{ds} \right) \left( \frac{D \dot{\gamma}^j}{ds} \right),
\]
where $\frac{D}{ds}$ is the covariant derivative along $\gamma$.

Under curvature flow, the geodesic curvature evolves by:
\[
\frac{d \kappa}{d \tau} = \Delta \kappa + \kappa^3 + \langle \text{Ric}(\dot{\gamma}), \dot{\gamma} \rangle,
\]
where $\Delta$ is the intrinsic Laplacian acting on scalar functions and Ric denotes the Ricci curvature tensor. This form appears in geometric flows that preserve curvature-aligned trajectories.

\subsection{Transport Interpretation: Curvature-Weighted Cost}

Let $p, q : \mathcal{M} \to \mathbb{R}$ be probability densities representing source and target patient manifolds. Define a cost function:
\[
c(z, z') = \| \nabla^2(z - z') \|^2,
\]
penalizing curvature-induced deformation.

Define the optimal transport problem with cost $c$:
\[
\text{OT}_c(p, q) = \inf_{\gamma \in \Pi(p, q)} \int_{\mathcal{M} \times \mathcal{M}} c(z, z') \, d\gamma(z, z'),
\]
where $\Pi(p, q)$ is the set of couplings with marginals $p$ and $q$.

If $T : \mathcal{M} \to \mathcal{M}$ is the Monge map pushing $p$ to $q$, the expected cost becomes:
\[
\mathbb{E}_{z \sim p} \left[ \| \nabla^2(z - T(z)) \|^2 \right].
\]

Then, the SOSM loss may be interpreted as:
\[
\mathcal{L}_{\text{SOSM}} \sim \text{OT}_{\text{curv}}(p, T_\#p),
\]
with curvature-weighted transport quantifying survival-consistent transformation across latent biological embeddings. In this view, minimizing $\mathcal{L}_{\text{SOSM}}$ aligns curvature-optimal deformation with survival-preserving geodesics.

\end{document}